%% file: arxiv.tex
\pdfoutput=1 % arXiv: process with pdfLaTeX
\documentclass[10pt,twocolumn,letterpaper]{article}

\usepackage{tikz}
\usepackage{float}
\usetikzlibrary{shapes,arrows}
\usetikzlibrary{calc}
\usetikzlibrary{fit}
\usetikzlibrary{positioning}
\usepackage{array}
\usepackage{placeins}
\usepackage{tabularx}
\usepackage{tcolorbox}
\tcbuselibrary{breakable}
\newtcolorbox{promptbox}[1]{
  colback=gray!3,
  colframe=black!50,
  boxrule=0.5pt,
  arc=1.5mm,
  title=#1,
  fonttitle=\bfseries\small,
  colbacktitle=gray!15,
  coltitle=black,
  top=3mm, bottom=3mm, left=4mm, right=4mm,
  fontupper=\small,
}
\newcommand{\promptrole}[1]{\textbf{#1:}\quad}
\usepackage[pagenumbers]{wacv} % To force page numbers, e.g. for an arXiv version


\colorlet{blue}{black} % arXiv: hide rebuttal revision highlights (\color{blue})

\definecolor{wacvblue}{rgb}{0.21,0.49,0.74}
\usepackage[pagebackref,breaklinks,colorlinks,allcolors=wacvblue]{hyperref}

\def\wacvPaperID{3102} % *** Enter the WACV Paper ID here
\def\confName{WACV}
\def\confYear{2027}

\newcommand{\modelname}{PathKGen}

\title{{\modelname}: Pathology Knowledge Assisted MLLM for Explainable Histopathology}

\author{
Nadun Rajapaksha$^{1,*}$ \quad
Sandun Herath$^{1,*}$ \quad
Devnith Wijesinghe$^{1,*}$ \quad
Isiri Amani Withanawasam$^{1,*}$ \\
Hasindri Watawana$^{2,3}$ \quad
Ranga Rodrigo$^{1,\dagger}$ \\
$^{1}$University of Moratuwa, Sri Lanka \quad
$^{2}$Idiap Research Institute, Switzerland \quad
$^{3}$EPFL, Switzerland\vspace{-0.8em} % arXiv: keeps Fig. 1 on page 1 despite the footnotes
}

\begin{document}
\maketitle
{\renewcommand{\thefootnote}{*}\footnotetext{These authors contributed equally to this work.}}
{\renewcommand{\thefootnote}{$\dagger$}\footnotetext{Correspondence: \href{mailto:ranga@uom.lk}{ranga@uom.lk}}}
    \begin{abstract}
    Whole Slide Image (WSI) analysis serves as the fundamental method for the clinical diagnosis of cancer in digital histopathology. Recent advances in Multimodal Large Language Models (MLLMs) have enabled state-of-the-art (SOTA) automated histopathology report generation. However, MLLMs reason purely through distributions encoded in their neural parameters, limiting the amount of knowledge and reasoning logic they can contain given the model scale. Their black-box reasoning poses another challenge to the human interpretation of intermediate reasoning. We address these issues by utilizing knowledge graphs to inject external structured knowledge for MLLMs to reason over, while making this reasoning component visible. Knowledge graphs are capable of partially externalizing the containment of medical knowledge, an area that remains underexplored in pathology report generation. We introduce PathKGen, a knowledge graph–assisted MLLM for histopathology report generation. PathKGen has two main branches: (1) a visual path that learns visual token inputs to the MLLM using pathology foundation models, and (2) a knowledge path that learns knowledge token inputs by extracting key morphological features of WSIs through querying a pathology-specific knowledge graph. A comprehensive evaluation on WSI-Bench dataset shows that PathKGen surpasses previous SOTA methods while revealing diagnostic reasoning as visually intuitive graphs. Due to its modular design, the proposed knowledge path is complementary to vision-aligned MLLMs and can be extended to other applications by curating domain-specific knowledge graphs to generate knowledge tokens. Source code is available at \href{https://github.com/nadunnr/PathKGen}{https://github.com/nadunnr/PathKGen}.
    \end{abstract}

\section{Introduction}
    \label{sec:intro}
    
    Histopathology serves as the foundation for cancer diagnosis, with Whole Slide Images (WSIs) capturing rich multi-scale morphological patterns at gigapixel resolution. Recent expansions in publicly available pathology datasets \cite{Weinstein2013TCGA,ikezogwo2023quilt,Liang2025wsillava} and availability of pathology foundation models \cite{lu2024conch,chen2024uni,ding2025titan} have popularized deep-learning based computational pathology workflows across a wide range of tasks such as zero-shot classification \cite{shi2024vilamil}, visual question answering (VQA) \cite{seyfioglu2024quiltlava}, and copilot assistant systems for pathology analysis \cite{chen2025slidechat,lu2024pathchat}. Recent adaptations of Multi-modal Large Language Models (MLLMs) for pathology have demonstrated strong performance in automated report generation \cite{seyfioglu2024quiltlava,guo2024histgen,Liang2025wsillava,lyu2025wsiagents}. However, existing models reason purely based on the distributions stored in the form of weights, which poses three fundamental limitations. 
    
    \begin{figure}[t]
    \centering
    \includegraphics[width=\columnwidth]{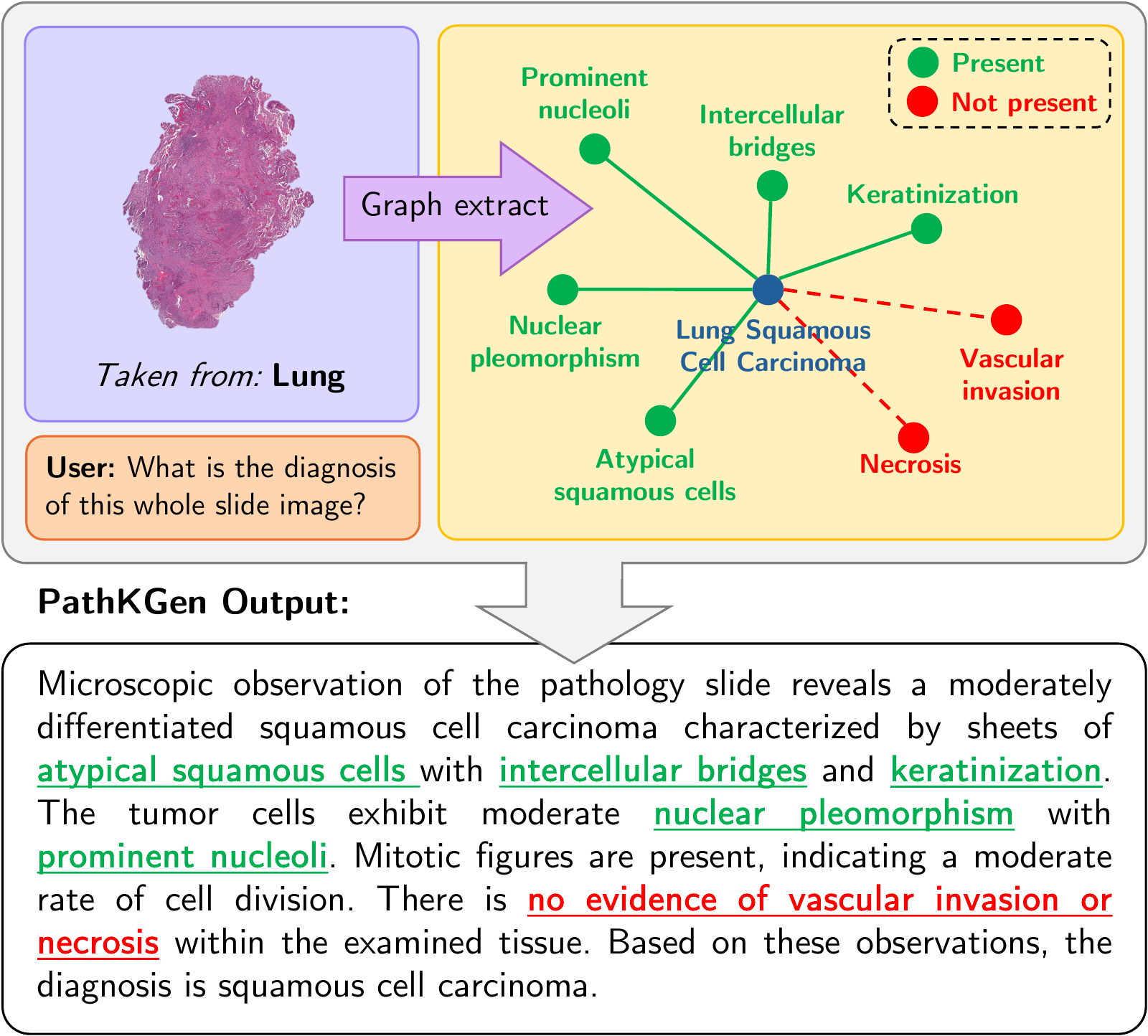}
    \caption{PathKGen is a vision-language assistant for generating histopathology reports while exposing the intermediate decision-making in a graphically intuitive, human readable format, thus improving explainability and encouraging clinical adoption.}
    \label{fig:teaser}
    \end{figure}

     The first limitation in recent slide-level MLLMs is their explainability. Prior works have trained on both closed-ended and open-ended VQA pairs which resulted in models that generate only final answers such as keywords or captions, without providing any underlying reasoning \cite{chen2025wsivqa, chen2025slidechat}. Their decision-making processes operate as black-box models, lacking transparency and human interpretability, which limits users from confidently trusting model outputs \cite{liu2025llmapplications,lu2024pathchat, Sivaroopan_2025_WACV}. WSI-LLaVA \cite{Liang2025wsillava} has attempted to improve explainability by incorporating morphological descriptions into the MLLM output. Yet its underlying reasoning remains largely implicit and parameter-bound. Greater explainability is therefore required to adopt trustworthy models into clinical settings. While medical experts naturally provide comprehensive reasoning alongside their diagnoses, the lack of explanatory depth in current models restricts physicians from confidently adopting and validating these WSI-level MLLMs~\cite{lopez2022llmpath, liu2025llmapplications}. 
     
    Secondly, MLLMs entirely handle the containment of domain knowledge and reasoning logic which binds the performance to the model scale. Because this knowledge is encoded directly within the model's weights, modifying the system with updated medical knowledge requires thorough fine-tuning and is computationally expensive. Thus, a knowledge reference technique analogous to that of a pathologist referring to a medical textbook is required for efficient adoption. In regard to this, medical Retrieval Augmented Generation (RAG) systems have proven to be versatile~\cite{xia2025mmedrag}, yet their contained knowledge is stored as vector representations making them impossible to comprehend and inconvenient to edit manually.
          
    The third challenge is that MLLMs are trained on pathologist-authored reports, which are inherently noisy and show reporting bias. WSI-Agents~\cite{lyu2025wsiagents} has utilized multi-agent systems to validate the internal reasoning logic of generated reports. Nevertheless, it relies on a combination of different MLLMs, thus remaining susceptible to noisy evidence propagation and increases overall parameter complexity, further compounding the aforementioned limitations.
    
    Knowledge graphs have emerged as a promising solution to outsourcing knowledge containment, gaining prominence as a widely employed data structure in medical deep learning systems \cite{jiang2024kare,li2024wikg,lou2025pathograph}. Their graphical representation is also capable of providing human interpretable visual intuition, which is ideal for explainability. However, their use for pathology knowledge containment and explainability is underexplored. Prior work~\cite{naseem2023kpathvqa} has used pathology knowledge graphs to guide the VQA tasks. Still, it directly encodes the extracted graph into dense embeddings, which nullifies human readability and violates transparency. As a comprehensive biomedical knowledge resource, the Unified Medical Language System (UMLS)~\cite{Bodenreider2004umls} integrates diverse medical concepts into a knowledge graph, that has been widely utilized across a broad range of clinical NLP tasks~\cite{ikezogwo2023quilt,zhou2024knowledge}. The integration of UMLS into medical domain pipelines capitalizes on a rigorously curated and standardized biomedical ontology, substantially reducing semantic noise, structural inconsistencies, and credibility risks associated with building a knowledge graph from the first principles.
    
    In this paper, we build on the success of MLLMs and the credible medical knowledge accumulated by the UMLS database, to propose \textbf{{\modelname}}, a knowledge graph guided MLLM which generates histopathology report text given a digital WSI as input. Our model tackles the limitations in existing MLLM based report generation models by revealing the intermediate reasoning through visually intuitive graphs, partially outsourcing the visual discrimination logic and knowledge containment, and providing a credible information signal to the MLLM. In our pipeline, slide-level features of the WSI are first extracted through pathology foundation models. We then extract characteristic morphological features of the histology present in the WSI, from a pathology specific knowledge graph derived from UMLS. The extracted subgraph is revealed to the user as shown in Fig. \ref{fig:teaser}. Finally, the MLLM is finetuned for accurate generation of report text by providing the slide feature tokens (visual tokens) and extracted morphological feature tokens (knowledge tokens) as input. We demonstrate that revealing intermediate decision making in a graphical format aids in improving explainability of the generated report.
    
    While our methodology emphasizes explainability, we also conduct studies on performance gains acquired through integrating a knowledge graph with an MLLM. We present standard natural language understanding (NLU) scores as well as the benchmark introduced in \cite{Liang2025wsillava} which are based on LLM-as-a-Judge techniques. Ablation studies show that the knowledge branch improves both model explainability and output quality at a given model scale, enabling smaller models to be on par with state-of-the-art (SOTA) models. We also outline the procedure for curating the pathology domain-specific knowledge graph from credible sources, which could be extended to other diagnostic fields. To the best of our knowledge, our method is the first to incorporate knowledge graphs for human interpretable pathology report generation.
    Our key contributions are summarized as follows:
    \begin{enumerate}
        \item We introduce a novel knowledge graph-assisted MLLM model architecture for pathology report generation, given the WSI.
    
        \item We outline a methodology for curating a pathology-specific knowledge graph derived from the UMLS database.
    
        \item We conduct comprehensive evaluation of our pipeline, and show that our methodology not only improves explainability, but also yields superior performance across all evaluated metrics, surpassing SOTA baselines.
    \end{enumerate}
    
    \begin{figure*}[!t]
        \centering
        \includegraphics[width=\textwidth]{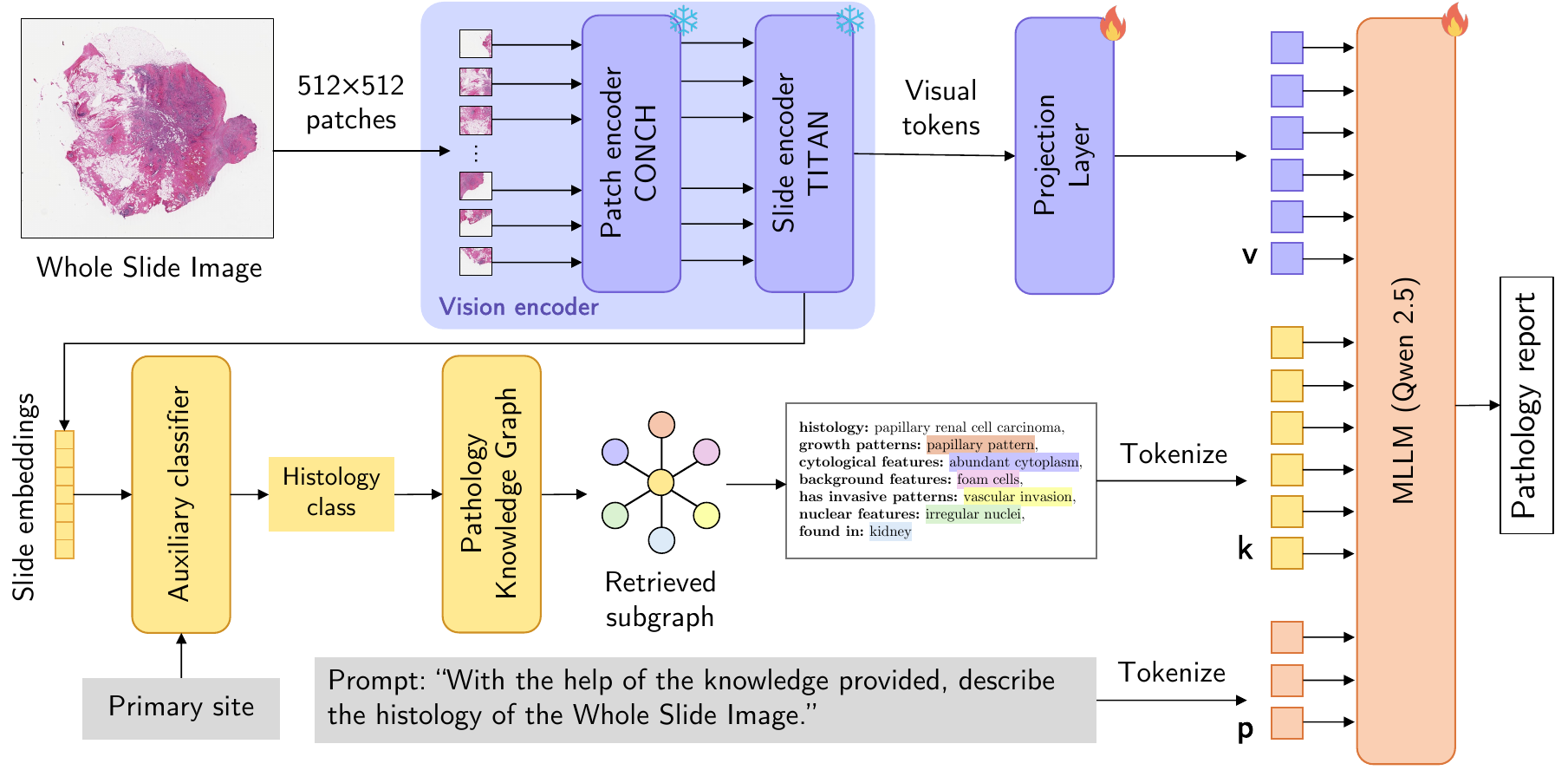}
        \caption{Overall pipeline architecture showing the connection of visual path (in blue) and knowledge path (in yellow). Slide features are extracted using CONCH and TITAN {\color{blue}(jointly serving as the vision encoder)}, and projected into the same semantic space of the MLLM. An auxiliary classifier predicts the histology in the WSI, which is subsequently used to query the knowledge graph and retrieve knowledge about morphological features. Visual tokens ($\mathbf{v}$), knowledge tokens ($\mathbf{k}$) and instruction prompt ($\mathbf{p}$) are given to the MLLM as input. See Sec. \ref{sec:method} for detailed methodology.} \label{fig1}
    \end{figure*}

\section{Related Work}

    \subsection{MLLMs for WSI Analysis}
        Recent use of MLLMs for WSI-level analysis and report generation demonstrate methods to overcome the gigapixel scale processing of WSIs. SOTA models adopt pathology specific vision foundation models as feature extractors. The standard routine follows the aggregation of patch level features extracted through patch encoders such as CONCH~\cite{lu2024conch}, and UNI~\cite{chen2024uni}, using transformer based slide encoder architectures like TITAN~\cite{ding2025titan}, PRISM~\cite{shaikovski2024prism}, and Madeleine~\cite{jaume2024madeleine}. The single slide-level representation tokens are linearly projected into the same semantic space of the MLLM following the LLaVA-style architecture. The visual tokens are combined with prompt tokens form the input to the MLLM, and is used to generate the report auto-regressively. SlideChat~\cite{chen2025slidechat}, Quilt-LLaVA~\cite{seyfioglu2024quiltlava}, and WSI-LLaVA~\cite{Liang2025wsillava} have adopted this standard framework.
    
    \subsection{Domain Knowledge Integrated WSI Analysis}
        Incorporating domain knowledge into WSI analysis has emerged as a promising direction for improving both predictive performance and interpretability. K-PathVQA~\cite{naseem2023kpathvqa} has used pathology knowledge graphs to guide the VQA task. It directly encodes the extracted graph into dense embeddings, which nullifies human readability and violates transparency. More recently, knowledge-enhanced WSI frameworks have explored explicit alignment between pathological concepts and tissue representations. For example, FOCUS~\cite{guo2025focus} aligns medical knowledge concepts with WSI regions to facilitate concept-aware feature compression and improve few-shot classification performance. 
    
        Similarly, MulGT~\cite{zhao2023mulgt} introduces task-specific knowledge injection and domain-guided graph pooling to capture diagnostically relevant tissue patterns across multiple pathology tasks. Despite these advances, existing methods primarily focus on improving classification or representation learning, while the integration of structured medical knowledge into pathology report generation remains unexplored. Our work addresses this gap by leveraging a pathology-specific knowledge graph to provide explicit, human-interpretable reasoning pathways for MLLM-based histopathology report generation.

\section{PathKGen}
    \label{sec:method}

    \begin{figure*}[t]
    \includegraphics[width=\textwidth]{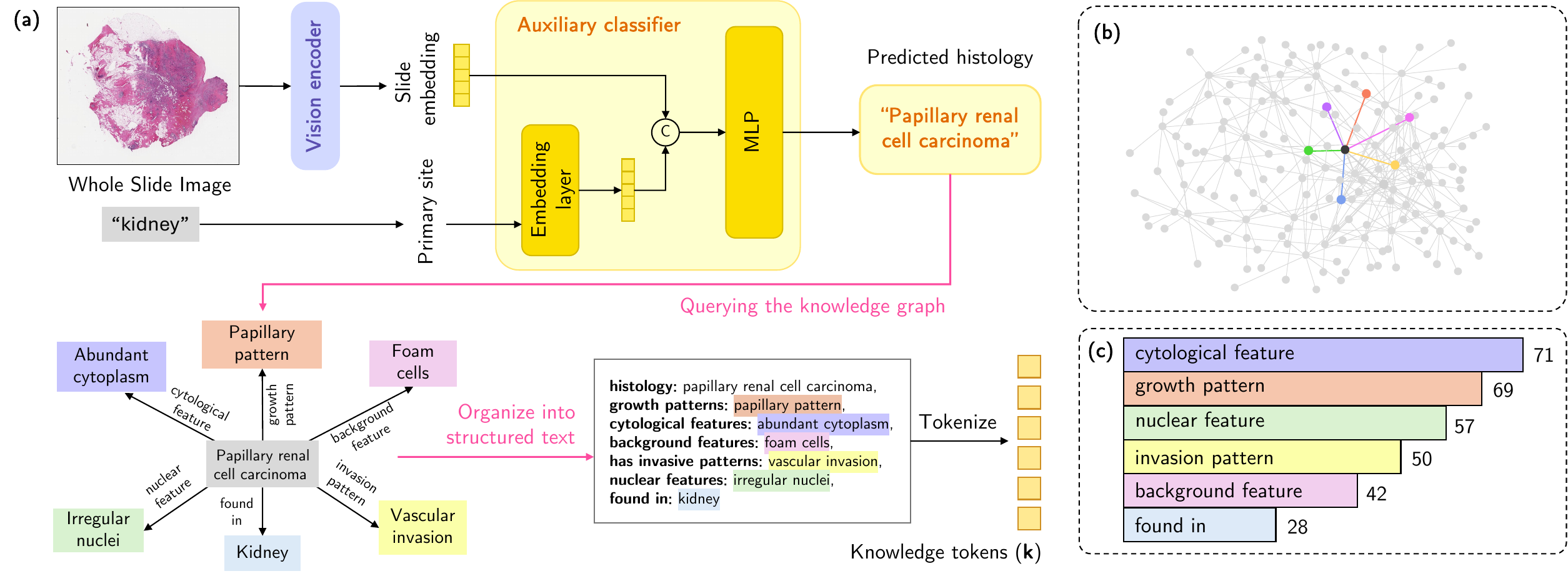}
    \centering
    \caption{{\color{blue}\textbf{(a)} Detailed working example of the how the subgraph is extracted, starting from the WSI. The auxiliary classifier predicts the histology class using slide embeddings and anatomical site, which acts as the anchor node for querying the pathology knowledge graph. This returns a fixed deterministic subgraph containing the canonical morphology profile of the predicted histology, which is then organized into structured text and tokenized. See Sec. \ref{sec:knowledge_path} for explanation of detailed working. \textbf{(b)} Comparison of the extracted subgraph with the base knowledge graph. \textbf{(c)} Distribution of the 6 categories of relations in our curated pathology knowledge graph.} } \label{fig:knowledge_path}
    \end{figure*}
    
    The complete pipeline of {\modelname} is illustrated in Fig. \ref{fig1}. Given an input WSI, the pipeline carries out two parallel pathways of processing to feed tokens to the MLLM and generate report text. The visual pathway first processes the WSI through pathology foundation models to extract slide features, which are then passed into a projection layer to get visual tokens. In parallel to the visual path, the knowledge path utilizes an auxiliary classifier module, which predicts the histology given the extracted slide features and the primary site of the WSI. Predicted histology is used as the anchor node to extract a subgraph from the curated knowledge graph by retrieving adjacent nodes. This subgraph is an intermediate result visible to the user in a graphical format (See Fig.~\ref{fig:teaser}). Finally, The extracted nodes and edge relations are then converted to structured text and appended to the MLLM input as knowledge tokens along with visual tokens to generate the report for a given instruction. The complete pipeline is trained in as an auto-regressive next-token prediction task~\cite{lauren2024idefics}, using the ground truth report text.
    
    \subsection{Visual Path}
    Leveraging the superior feature representation capabilities of pathology foundation models, we adopt CONCHv1.5 vision encoder \cite{lu2024conch} to extract 768-dimensional patch features from WSIs. To aggregate the patch features into a single slide-level representation, we use the TITAN slide encoder \cite{ding2025titan}, which outputs slide tokens $\mathbf{x}$ of size 128 $\times$ 768. Following the auto-regressive LLaVA architecture \cite{liu2023visualinstructiontuning}, the slide tokens $\mathbf{x}$ are projected into the semantic space of the MLLM by passing through a modality projection layer, which consists of a two-layer Multi-Layer Perceptron (MLP) block. The resulting projected visual tokens $\mathbf{v}$ are combined with the prompt and fed into the MLLM. Simultaneously, to train the auxiliary classifier, slide tokens are  pooled through an attention mechanism \cite{ding2025titan} and extracted as 768-dimensional slide embeddings denoted by $\mathbf{x_p}$.

    \subsection{Knowledge Path}
    \label{sec:knowledge_path}
    {\color{blue} The fundamental task of the knowledge path is to predict the histology in the WSI, retrieve the canonical morphological profile of that class from the knowledge graph and feed it into the MLLM in a structured format.  This enhances the MLLM’s ability to learn discriminative histological features while making its decisions human-readable and partially externalizes knowledge containment to the knowledge path.}\\

    \noindent \textbf{Auxiliary Classifier.} For weakly supervised training of the auxiliary classifier, we first prepare histology class label $h$ and primary site label for each WSI using the metadata available through The Cancer Genome Atlas (TCGA) \cite{Weinstein2013TCGA} portal. To aid the classifier differentiate between similar carcinomas arising at different organs, we also feed primary site as an input in the form of a learnable embedding, $\mathbf{s}$. This approach mimics the clinical workflow, where pathologists narrow down potential histologies based on the primary site. We notice that this support from the available metadata enables the classifier to achieve a higher classification accuracy, compared to without. We then train a lightweight MLP head to predict the histology class $\hat{h}$ conditioned on the concatenated slide embedding and site embedding, as shown in Equation \ref{eq:aux_classifier}.

    \begin{equation}
    \label{eq:aux_classifier}
        \hat{h} = \arg\max_{h \in H} P(h\mid \mathbf{x_p};\mathbf{s})
    \end{equation}
    
    \begin{figure*}[t]
    \includegraphics[width=\textwidth]{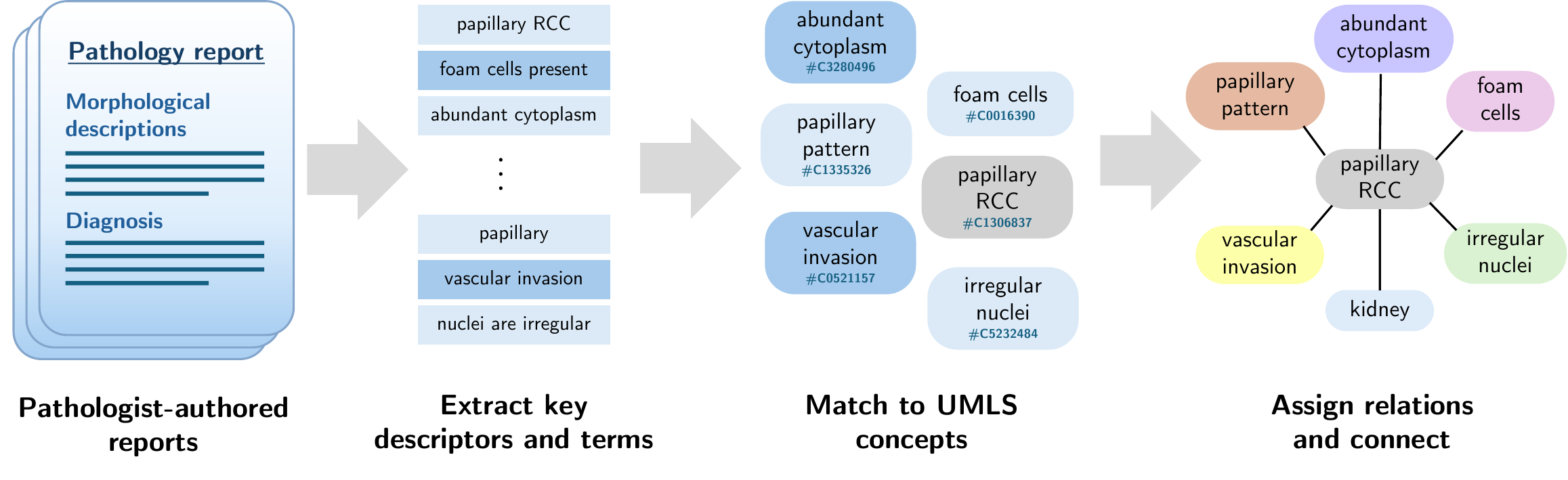}
    \centering
    \caption{Knowledge graph curation pipeline. Morphological features are extracted from ground-truth histopathology reports for each histology class and mapped to UMLS concepts using a biomedical concept alignment model. Then semantic relations between histology classes and morphological concepts are assigned based on UMLS semantic types. The resulting knowledge graph is subsequently reviewed and verified by expert pathologists. See Sec.~\ref{sec:kg curation} for more detailed description.}
    \label{curation}
    \end{figure*}
    where $H$ defines the set of all possible histology classes and $P(\cdot)$ denotes the probability function.
    
    \input{tables/table5_classifier}
    
    We train the auxiliary classifier for 31 histology classes (including variants) with the help of 52 primary sites. The size of the learnable site embedding is chosen to be 16-dimensional. The auxiliary classifier is trained separately, prior to training the MLLM, under cross-entropy loss. It achieves an overall accuracy of 97.52\% and a macro-averaged F1 score of 0.97, demonstrating balanced performance across all histology classes.
    {\color{blue}
    Table~\ref{tab:histology_classification_ablation} shows that primary site alone achieves a Macro-F1 of 0.6941, while slide embedding alone achieves 0.8754. Combining both improves Macro-F1 to 0.9711, indicating that primary-site information provides complementary discriminative information to the visual slide representation. Per-class scores and confusion matrix (Appendix~\ref{sec:supp_aux_classifier}) confirm high accuracy under prevailing class imbalance.\\}
    
    \noindent \textbf{Querying the Knowledge Graph.} We query for knowledge from the curated knowledge graph, derived from UMLS, using the predicted histology of each WSI as the anchor node. Denoting the knowledge graph as $G = (V, E)$ and anchor node $n_{\hat{h}}\in V$, we retrieve the \textit{one-hop induced neighborhood subgraph} of $n_{\hat{h}}$, which is a tree denoted by $G_{\hat{h}} \subseteq G$. The leaf nodes of $G_{\hat{h}}$ are morphological features of the histology $\hat{h}$, while the edges of $G_{\hat{h}}$ represent the semantic relations between the histology and each morphological feature. These semantic relations fall into 6 categories as shown in Fig. \ref{fig:knowledge_path}, namely,
    ~\textit{found in, growth pattern, cytological feature, nuclear feature, background feature} and \textit{invasion pattern}. 
    This categorization of links helps to organize the morphological features into a structured text format. This text is then tokenized and fed into the MLLM as knowledge input $\mathbf{k}$. We adopt the text structure shown in Fig. \ref{fig:knowledge_path}, as it is highly compatible with the chosen MLLM. Furthermore, the knowledge retrieved from $G$ can be shown as a visually intuitive graph. 
    
    {\color{blue} The retrieved subgraph is class-conditional and deterministic: two slides assigned the same histology receive identical knowledge tokens $\mathbf{k}$. Slide-specific content enters the report \textit{only} through the visual tokens $\mathbf{v}$, against which the MLLM confirms, negates, or omits each retrieved candidate morphology during generation. This mirrors the clinical practice, where a pathologist narrows to a likely entity, recalls the features that define it, and then checks the slide for each one.}
    
    \begin{figure*}[t]
    \includegraphics[width=\textwidth]{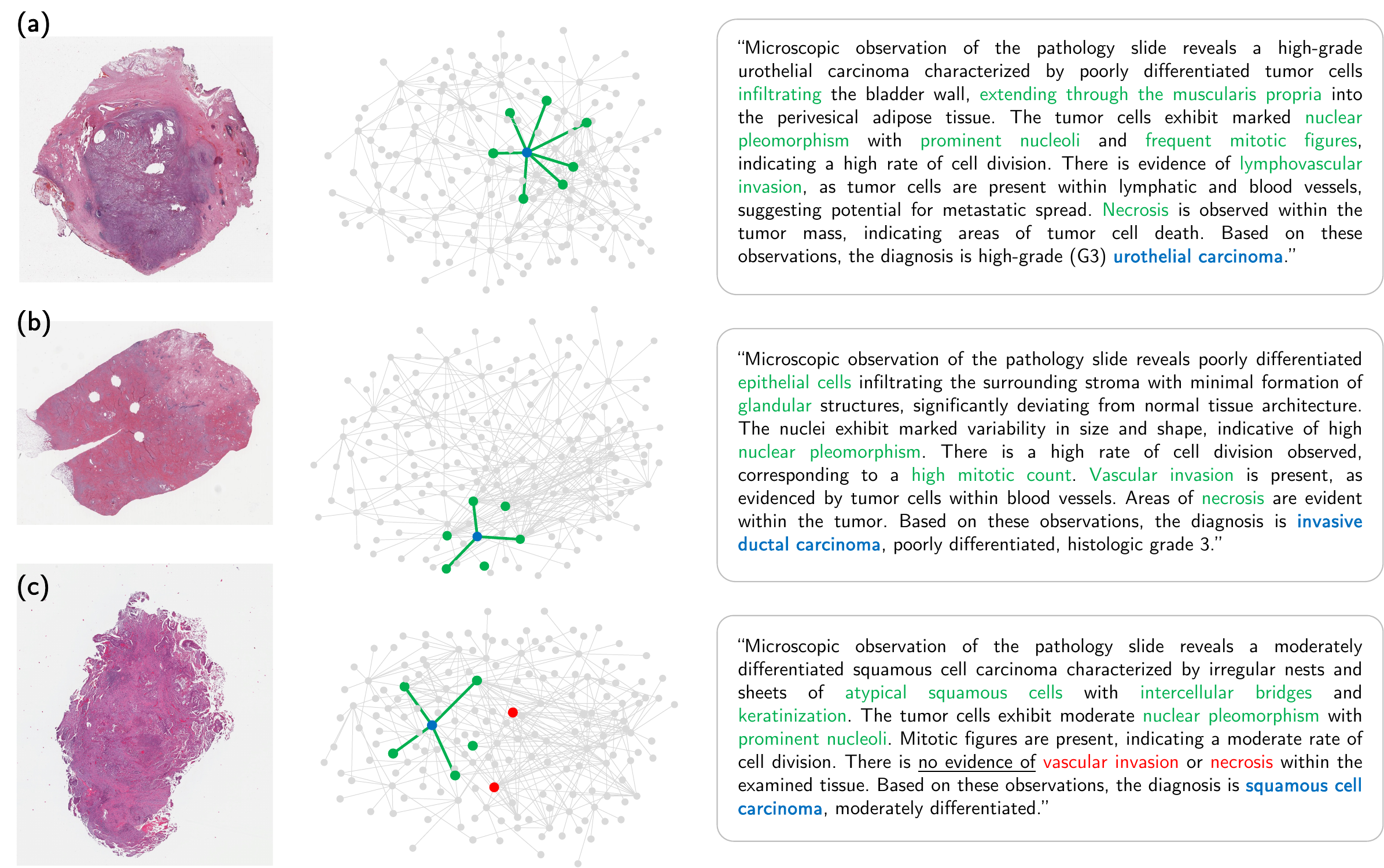}
    \centering
    \caption{Qualitative visualization of sample model predictions, highlighting the extracted feature nodes (in comparison to the entire knowledge graph) which are explicitly reflected in the generated report. Predicted carcinoma is highlighted in blue. Features present in the WSI are highlighted in green, while features with no evidence are highlighted in red. See Sec.~\ref{sec: qualitative results interpretation} for the interpretation. Node labels are omitted for clarity and could be toggled on using graph visualization software. {\color{blue} The color coding is a post-hoc formatting processed using rule-based text matching. Additional examples with node labels are provided in Appendix~\ref{sec:supp_qualitative}.}} \label{fig:qualitative_examples}
    \end{figure*}
    
    \subsection{Language Model} We adopt Qwen2.5-7B-Instruct model as our MLLM decoder due to its capability of working with multi-modal data~\cite{qwen2.52024qwen}. However, Vicuna~\cite{zheng2023llmjudge} and LLama3~\cite{touvron2023llama} have also been proven to be capable for this task~\cite{chen2025slidechat}. Inspired from LLaVA-style training~\cite{liu2023visualinstructiontuning}, the training is performed in two stages. First, we train only the projection layer while keeping the MLLM backbone frozen for the purpose of semantic space alignment. Subsequently, we load the unfrozen MLLM using QLoRA~\cite{dettmers2023qlora} for parameter efficient visual instruction tuning and train both the projection layer and MLLM as an auto-regressive next token prediction task. The input to the model is a single unified prompt $\mathcal{P} = [\, \mathbf{v} \,;\, \mathbf{k} \,;\, \mathbf{p} \,]$ consisting of the visual tokens $\mathbf{v}$, knowledge tokens $\mathbf{k}$, and the instruction prompt $\mathbf{p}$. The confidence of the auxiliary classifier is explicitly stated in the instruction prompt $\mathbf{p}$. This prevents the MLLM from solely relying on the classifier output. The model is trained to predict the output $\hat{y}$.
    
    The unified prompt $\mathcal{P}$ is masked to create the training labels, such that when predicting the $\hat{y}$, visual and knowledge tokens are attended. The training objective of cross entropy loss for the next token given $\mathcal{P}$ is augmented with an adaptive weighting which prioritizes challenging examples (examples which score low in NLU metrics) by increasing their loss contribution while down-weighting trivial examples~\cite{chang2025bleuberi}.

    \subsection{Knowledge Graph Curation}
    \label{sec:kg curation}

    The knowledge graph was curated with data derived from the UMLS Metathesaurus \cite{Bodenreider2004umls}. We utilize GPT-4o~\cite{gpt4o} to extract crucial morphological features for each histology class from ground truth reports of WSI-Bench~\cite{Liang2025wsillava}, and link each terminology to UMLS concepts using SapBERT~\cite{liu2021sapbert}, which is a pre-trained framework that aligns biomedical text into UMLS concepts. We then filter the matched concepts by the frequency of occurrence to remove matches which are likely to be noise. Finally, by considering the UMLS semantic type and the path connecting each morphological feature to the histology class, we use GPT-4o~\cite{gpt4o} to categorize each semantic relation into one of the 6 relations. Consequently, we verify the integrity of the curated knowledge graph by pathologists.
    
    The curated knowledge graph has 177 nodes and 317 edges. Nodes consist of histologies and morphological features, while edges are of 6 categories as shown in Fig. \ref{fig:knowledge_path}. Statistical analysis shows a sparse yet cohesive graph, characterized by graph density of 0.01 and average node out-degree of 1.79. Fig. \ref{fig:knowledge_path} $\mathbf{(b)}$ also depicts the distribution of edges across the 6 categories of relations in the knowledge graph.
    
    \input{tables/table1}

\input{tables/table2_kg_ablation}
    \input{tables/table4_param_effec}
    
    \subsection{Dataset and Training.}
    We use the WSI-Bench dataset \cite{Liang2025wsillava}, which was curated from TCGA \cite{Weinstein2013TCGA}, as our primary dataset for the report generation pipeline. The report generation sub-task of this dataset includes 9749 training pairs and 208 test pairs, each consisting of a WSI and its corresponding pathology report. From that, a total of 9452 slides were used after removing slides which do not include microns per pixel value. {\color{blue} For weakly supervised training of the auxiliary classifier, we derive histology labels from WSI reports using rule-based text processing and obtain primary-site metadata from TCGA. Extracted labels were cross-validated against labels available from the TCGA portal. The result is a supplementary dataset containing WSIs paired with primary site inputs and the histology classes as ground truth labels.}

\section{Experiments}
\label{sec:experiments}

    \subsection{Quantitative Results on WSI-Bench Dataset.}
    \label{sec:quantitative results}

    We evaluate our method against existing WSI MLLMs under the report generation task of the WSI-Bench dataset \cite{Liang2025wsillava}. As presented in Table \ref{tab1}, {\modelname} outperforms existing MLLMs including MI-Gen~\cite{chen2024wsicaption}, HistGen~\cite{guo2024histgen}, WSI-LLaVA~\cite{Liang2025wsillava} and WSI-Agents~\cite{lyu2025wsiagents}, in multiple metrics. Specifically, our model achieves notable relative improvements across five standard NLU scores: BLEU-1 by 6.67\%, BLEU-2 by 8.66\%, BLEU-3 by 10.10\%, BLEU-4 by 8.75\%, ROUGE-L by 5.92\%, and METEOR by 2.75\%.
    
    Furthermore, as histopathology report generation is a domain characterized by complex and highly nuanced terminology, standard NLU metrics alone are inadequate to comprehensively assess report quality. To rigorously evaluate performance in this context, we employ an LLM-as-a-Judge metric: WSI-Precision introduced in \cite{Liang2025wsillava}, which evaluates the accuracy of generated reports by verifying the model predictions against the ground truth claims. Our approach achieves a WSI-Precision of 0.480, outperforming the SOTA by 9.09\%. These results demonstrate that {\modelname} surpasses existing methods by significant margins.
    
    \subsection{Evaluation on Semantic Alignment}
    
    We evaluated our model outputs using BERTScore, a deep learning based metric that measures the semantic alignment between the ground truth and the generated text \cite{zhang2019bertscore}. {\modelname} achieves a BERTScore F1 of 0.9081 with RoBERTa-large model~\cite{liu2019roberta} and 0.8836 with DistilBERT model~\cite{Sanh2019distillbert}. Hallucinations or omissions of critical medical terms would be reflected by a significant decrease in BERTScore. Our consistently high BERTScore indicates strong semantic alignment with the ground truth and minimal factual distortion. In comparison, WSI-LLaVA achieves a BERTScore of 0.9068 for RoBERTa-large. These findings suggest that the MLLM is robust to uncertainty in the classifier outputs, with no observable degradation in report quality. 
    
    \subsection{Qualitative Results Interpretation}
    \label{sec: qualitative results interpretation}

    Fig. \ref{fig:qualitative_examples} shows selected outputs from the model for qualitative analysis. We observe that the key morphological descriptors of the carcinoma illustrated by the subgraph are reflected in the generated report. In  Fig.~\ref{fig:qualitative_examples} $\mathbf{(a)}$, the subgraph features and descriptors in the report show high overlap, meaning the MLLM faithfully adapts to the information provided by the knowledge path.
    
    As illustrated in Fig.~\ref{fig:qualitative_examples} $\mathbf{(b)}$, the MLLM also learns novel connections absent from the knowledge graph (isolated green nodes), demonstrating the \textit{partial} externalization of its internal knowledge. Since trivial characteristics are consistently supported by the knowledge path, the model parameters instead learn more complex relationships between visual descriptors and carcinomas. Furthermore, the MLLM learns negation rules, whereby the absence of specific morphological features serves as positive diagnostic evidence for certain histologies. This is confirmed by isolated red nodes in Fig.~\ref{fig:qualitative_examples} $\mathbf{(c)}$, indicating that absent features remain disconnected from the corresponding histology node.

\input{tables/table3_counterfactual}
    
    \subsection{Ablation Studies}
    \label{sec: ablation_studies}

    {\color{blue}
    We ablate the contribution of the knowledge path at the 7B parameter scale. Compared with the vision-only baseline, \modelname{} improves BLEU-4 by 2.47\%, METEOR by 2.52\%, and WSI-Precision by 3.03\% as shown in Table~\ref{tab:knowledge_ablation}. To determine whether these gains are merely from conditioning on the auxiliary classifier output or primary site metadata, we additionally evaluate variants where the knowledge pathway receives only the predicted histology or only the primary site. \modelname{} consistently outperforms both baselines, indicating that the richer structured knowledge, including the subgraph relations and associated morphological features, provides complementary information beyond histology or site conditioning alone.}

    {\color{blue}
    Further, the counterfactual interventions in Table~\ref{tab:counterfactual_kg} show that oracle knowledge slightly improves performance, whereas random incorrect histology anchors substantially degrade WSI-Precision, confirming that the model meaningfully utilizes the supplied knowledge.
    }
    
    To understand the amount of knowledge externalization handled by the knowledge path, we compare three model sizes with the same knowledge path architecture (Table~\ref{tab:parameter_efficiency}).  Lower-parameter models achieve performance comparable to 7B SOTA, supporting that visual discrimination logic and knowledge containment of the MLLM have become partially externalized to the knowledge path. Inevitably, scaling the model size gives a greater observable performance gain than from the knowledge path. However, partial externalization of domain knowledge not only improves the explainability but also increases the model performance at a given scale, by providing an external guidance to the MLLM.

\section{Conclusion}
    We presented a knowledge graph-assisted MLLM for pathology report generation    with human-readable intermediate results, demonstrating a methodology for curating a credible pathology knowledge graph with outsourced knowledge containment and integrating it into a standard MLLM architecture. To the best of our knowledge, this is the first method to do so. Experiments on a benchmark dataset showed superior performance over existing models, while ablations confirm the effectiveness of the knowledge-addition methodology. We leave generalizing this method to other diagnostic routines as future work.
{
    \small
    \bibliographystyle{ieeenat_fullname}
    \bibliography{main}
}

% ===================== arXiv: Supplementary Material (from supplementry.tex) =====================
\clearpage
\onecolumn
\appendix
\begin{center}
{\Large\bfseries Supplementary Material\par}
\vspace{1.0em}
\end{center}

\section{Analysis of Auxiliary Classifier.}\label{sec:supp_aux_classifier}

\begin{table}[H]
\centering
\caption{
Per-class histology classification metrics for the auxiliary classifier evaluated on the test set.
}
\label{tab:per_class_histology}

\begin{tabular}{@{} p{8.2cm} 
    >{\raggedleft\arraybackslash}p{1.5cm} 
    >{\raggedleft\arraybackslash}p{1.5cm} 
    >{\raggedleft\arraybackslash}p{1.5cm} 
    >{\raggedleft\arraybackslash}p{1.5cm} @{}}
\toprule
Class & Precision & Recall & F1-score & Support \\
\midrule
Adrenocortical Carcinoma
& 1.00 & 1.00 & 1.00 & 5 \\

Bladder Urothelial Carcinoma
& 1.00 & 0.89 & 0.94 & 9 \\

Brain Lower Grade Glioma
& 1.00 & 1.00 & 1.00 & 15 \\

Breast Invasive Carcinoma
& 1.00 & 0.96 & 0.98 & 23 \\

Cervical Squamous Cell Carcinoma
& 1.00 & 1.00 & 1.00 & 2 \\

Cholangiocarcinoma
& 1.00 & 1.00 & 1.00 & 2 \\

Colon Adenocarcinoma
& 1.00 & 1.00 & 1.00 & 8 \\

Endocervical Adenocarcinoma
& 1.00 & 1.00 & 1.00 & 2 \\

Esophageal Carcinoma
& 1.00 & 1.00 & 1.00 & 3 \\

Head and Neck Squamous Cell Carcinoma
& 1.00 & 1.00 & 1.00 & 9 \\

Kidney Chromophobe
& 1.00 & 1.00 & 1.00 & 2 \\

Kidney Renal Clear Cell Carcinoma
& 1.00 & 1.00 & 1.00 & 11 \\

Kidney Renal Papillary Cell Carcinoma
& 1.00 & 1.00 & 1.00 & 6 \\

Liver Hepatocellular Carcinoma
& 1.00 & 1.00 & 1.00 & 7 \\

Lung Adenocarcinoma
& 0.90 & 0.90 & 0.90 & 10 \\

Lung Squamous Cell Carcinoma
& 0.90 & 0.90 & 0.90 & 10 \\

Lymphoid Neoplasm Diffuse Large B-cell Lymphoma
& 1.00 & 1.00 & 1.00 & 2 \\

Mesothelioma
& 1.00 & 1.00 & 1.00 & 2 \\

Ovarian Serous Cystadenocarcinoma
& 0.67 & 1.00 & 0.80 & 2 \\

Pheochromocytoma
& 1.00 & 1.00 & 1.00 & 4 \\

Prostate Adenocarcinoma
& 1.00 & 1.00 & 1.00 & 9 \\

Rectum Adenocarcinoma
& 1.00 & 1.00 & 1.00 & 4 \\

Sarcoma
& 1.00 & 1.00 & 1.00 & 10 \\

Skin Cutaneous Melanoma
& 1.00 & 1.00 & 1.00 & 9 \\

Stomach Adenocarcinoma
& 1.00 & 1.00 & 1.00 & 8 \\

Thymoma
& 1.00 & 1.00 & 1.00 & 4 \\

Thyroid Carcinoma
& 1.00 & 1.00 & 1.00 & 10 \\

Uterine Carcinosarcoma
& 0.50 & 1.00 & 0.67 & 2 \\

Uterine Corpus Endometrial Carcinoma
& 1.00 & 0.90 & 0.95 & 10 \\

Uveal Melanoma
& 1.00 & 1.00 & 1.00 & 2 \\

\midrule
Accuracy
& \multicolumn{3}{r}{0.98} & 202 \\

Macro Avg.
& 0.97 & 0.98 & 0.97 & 202 \\

Weighted Avg.
& 0.98 & 0.98 & 0.98 & 202 \\
\bottomrule
\end{tabular}
\end{table}

\begin{figure}[htbp]
    \centering
    \includegraphics[width=0.7\textwidth]{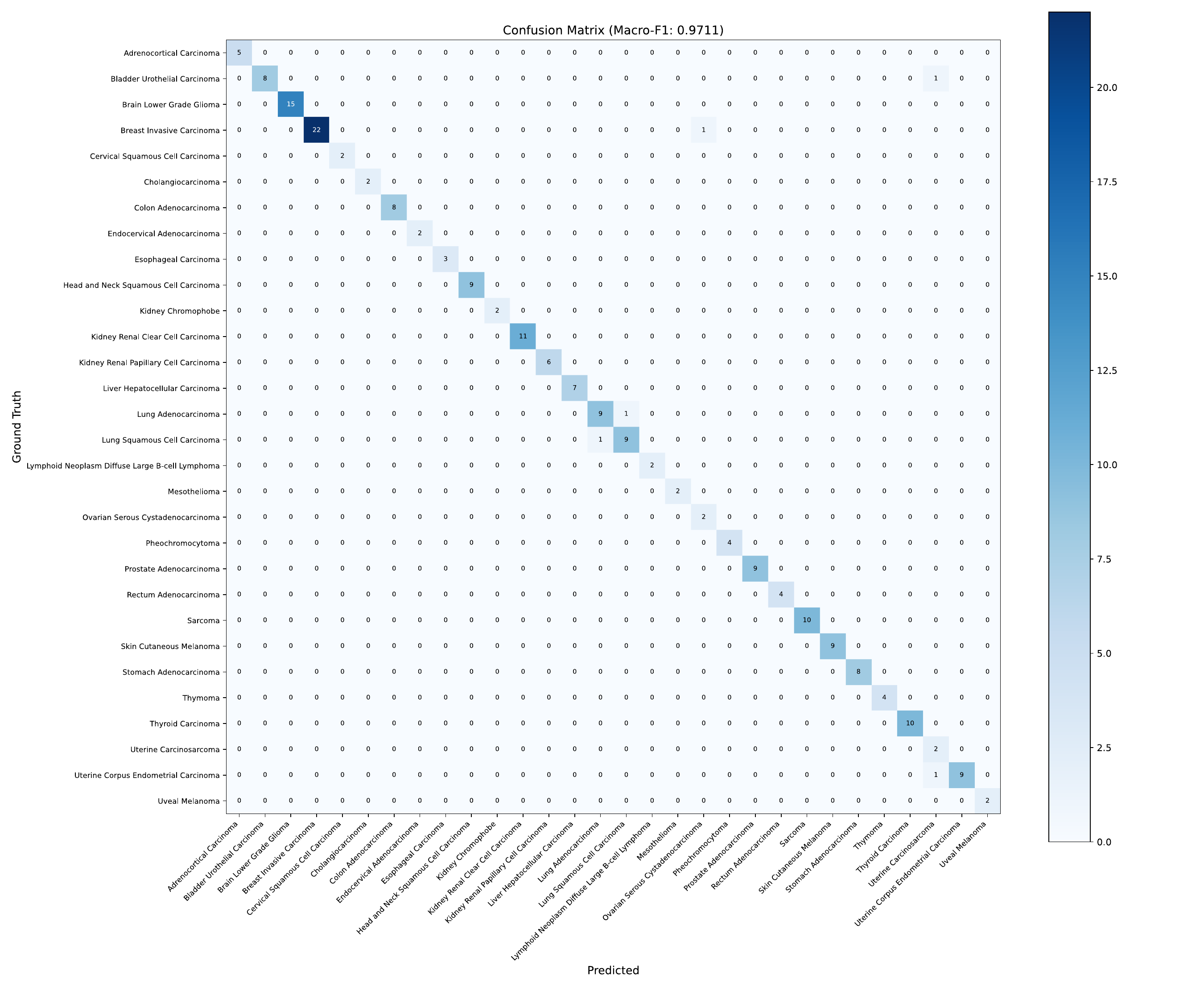} % arXiv: SVG pre-converted to PDF 
    \caption{Confusion matrix for the auxiliary classifier (Sec. 3.2), evaluated on the test set}
    \label{fig:svg_example}
\end{figure}

\section{Training Details.}\label{sec:training_details}

After segmenting the tissue region the whole slide images, are tiled into $512\times512$ patches at $20\times$ magnification discarding the patches which does not include tissue regions. Then all the filtered patches are encoded with the CONCH v1.5 patch encoder. Patch level features are aggregated into slide representations using the TITAN slide encoder.

As mentioned in the paper training proceeds in two stages. In stage 1, only the projection layer is trained keeping the LLM backbone frozen, for 8 epochs with effective batch size 64, using Adam optimizer with learning rate $5\times 10^{-4}$ and a weight decay of $0.01$. In Stage 2, both projection layer and the LLM backbone are fine tuned with QLoRA, for one epoch with effective batch size 128, using AdamW optimizer with a learning rate of $1\times10^{-4}$ for the QLoRA-adapted LLM parameters, $1\times10^{-5}$ for the projection layer and a weight decay of $0.01$. LoRA adapters use rank $r=128$ and $\alpha=256$, with dropout $0.02$, applied to the attention and MLP projection matrices. Additionally, we use 42 as the random seed for functions from every python class.

\subsection{Adaptive Loss Formula.}
The training objective is per-token autoregressive cross entropy, computed only over the report segment of each sequence, with label smoothing $\epsilon = 0.1$ applied:

\begin{equation}
\mathcal{L}_{\text{CE}} = -\frac{1}{|T|}\sum_{t \in T} \log p_\theta(y_t \mid x_{<t})
\label{eq:ce_loss}
\end{equation}

\noindent where $T$ is the set of unmasked token positions.

At periodic steps, the loss is reweighted using a BLEU-4-based hard example signal: the model greedily generates a report for each sample $i$, scores it against the reference ($b_i$), and computes a weight relative to a running baseline $\beta$:

\begin{equation}
w_i = 0.5 + \frac{1}{1+\exp\big(5(b_i-\beta)\big)} \in (0.5,\ 1.5), \qquad
\beta \leftarrow 0.9\,\beta + 0.1\,\bar{b}
\label{eq:bleu_weight}
\end{equation}

\noindent with $\beta$ initialized to $0.2$ and $\bar{b}$ the mean BLEU-4 over the mined micro-batch.

% \subsubsection{Validation Protocol}f
%     Validation loss and NLU/WSI-Precision metrics are computed every epoch on the held-out split. Report generation during evaluation uses beam search with 4 beams, length penalty $1.0$, and a no-repeat n-gram size of $3$.

% \subsubsection{Hardware and Random Seed}
% All experiments use random seed 42 (applied to \texttt{torch}, \texttt{numpy}, and \texttt{random}). Training was performed on 1 NVIDIA RTX 3090 GPU.

\section{Multiple Run Statistics.}
\input{tables/conf_int}

\section{Prompt Template.}
\label{sec:prompt_templates}

All models are trained and prompted using a format with three roles (\texttt{system}, \texttt{user}, \texttt{assistant}). The visual features from the TITAN slide encoder are inserted at the position of a pair of reserved markers, \texttt{<image\_start>} and \texttt{<image\_end>}, which are added to the tokenizer as special tokens. During training, only the tokens belonging to the \texttt{assistant} turn (i.e., the report or answer) are supervised; every token up to and including \texttt{<|im\_start|>assistant} is masked with a label of $-100$ so that the model learns to \emph{condition on} the knowledge and question rather than to reconstruct them (Sec.~\ref{sec:training_details}).

\begin{figure}[H]
\centering
\begin{promptbox}{Generation Prompt}
\promptrole{System} You are a pathology assistant specialized in analyzing whole slide images of tissue samples. Provide detailed, accurate diagnostic observations based on microscopic features.\\

\promptrole{User}\texttt{<image\_start><image\_end>} \\
Generic knowledge regarding histology type (97.52\% accurate). The site is exact. \\
\texttt{\{knowledge\}} \\
Use to help, but verify visually.

\texttt{\{question\}}\\

\promptrole{Assistant}\emph{\{answer\}}
\end{promptbox}
\caption{Generation prompt template with knowledge guidance. During training, the pathologist-written report is given as the answer in the assistant turn; at inference, autoregressive next token generation begins from the assistant turn.}
\label{fig:generation_prompt}
\end{figure}

\section{Qualitative Analysis.}\label{sec:supp_qualitative}
\begin{figure}[H]
    \centering
    \includegraphics[
    width=0.9\linewidth]{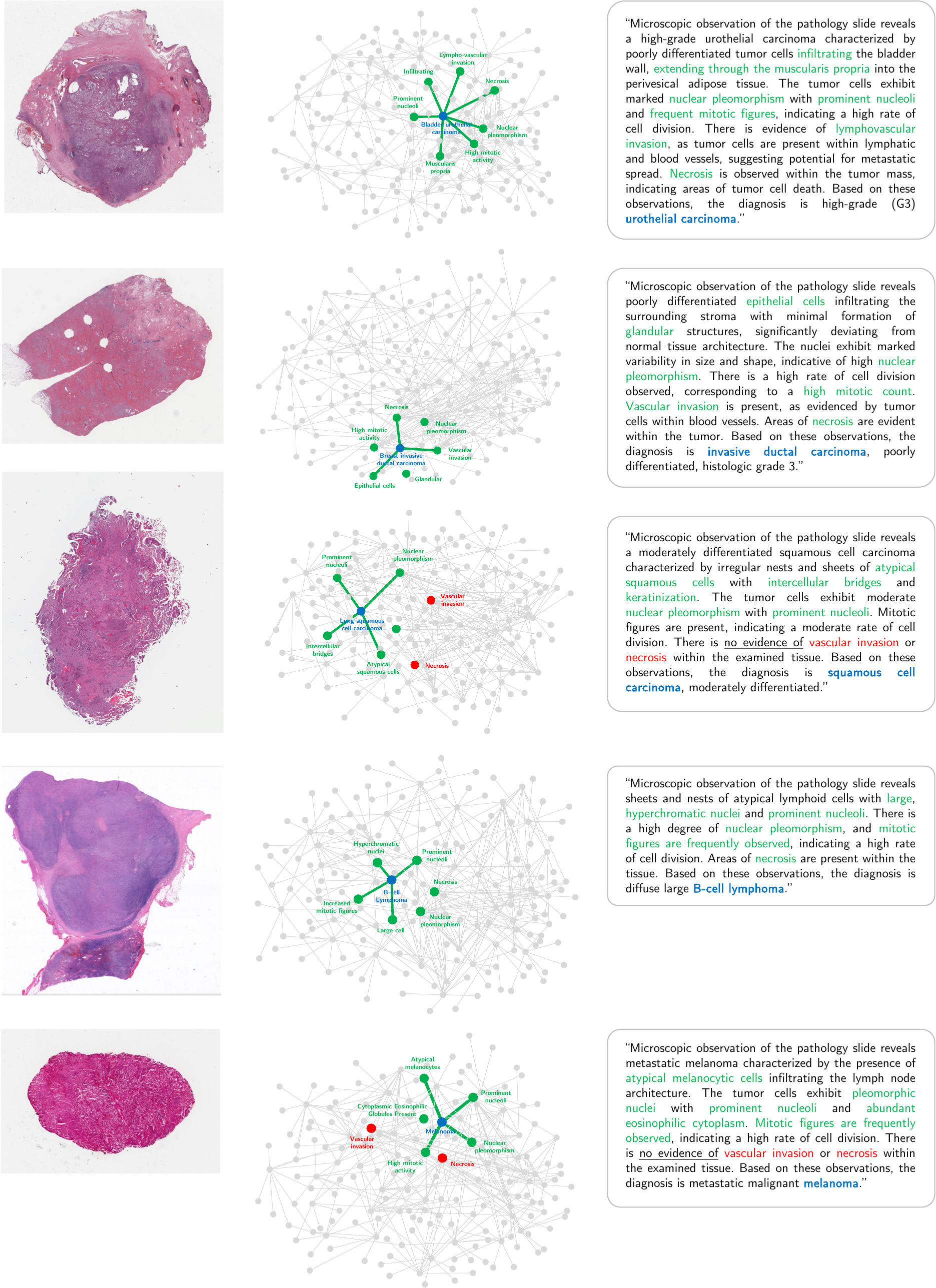}
    \caption{Additional qualitative examples containing triplets of Whole Slide Image (WSI), activated graph nodes, and generated report. Green: evidence present in the WSI, red: evidence absent in the WSI, blue: predicted histology.}
    \label{fig:full_examples}
\end{figure}

Figure \ref{fig:full_examples} shows additional qualitative visualization of sample model predictions with extracted feature nodes highlighted. Features present in the WSI are highlighted in green, while features with no evidence are highlighted in red. This formatting is done through rule-based text processing for post-hoc intuitive display of present and absent features. Tables \ref{tab:example_1},  \ref{tab:example_2} and \ref{tab:example_3} compares generated reports from WSI-LLaVA and {\modelname} against the ground truth sample. As evident qualitatively, {\modelname} captures more semantics in the report than WSI-LLaVA.

\newpage

\input{tables/idx52}
\input{tables/idx57}
\input{tables/idx138}

\newpage
\begin{figure}[H]
    \centering
    \includegraphics[width=0.9\linewidth]{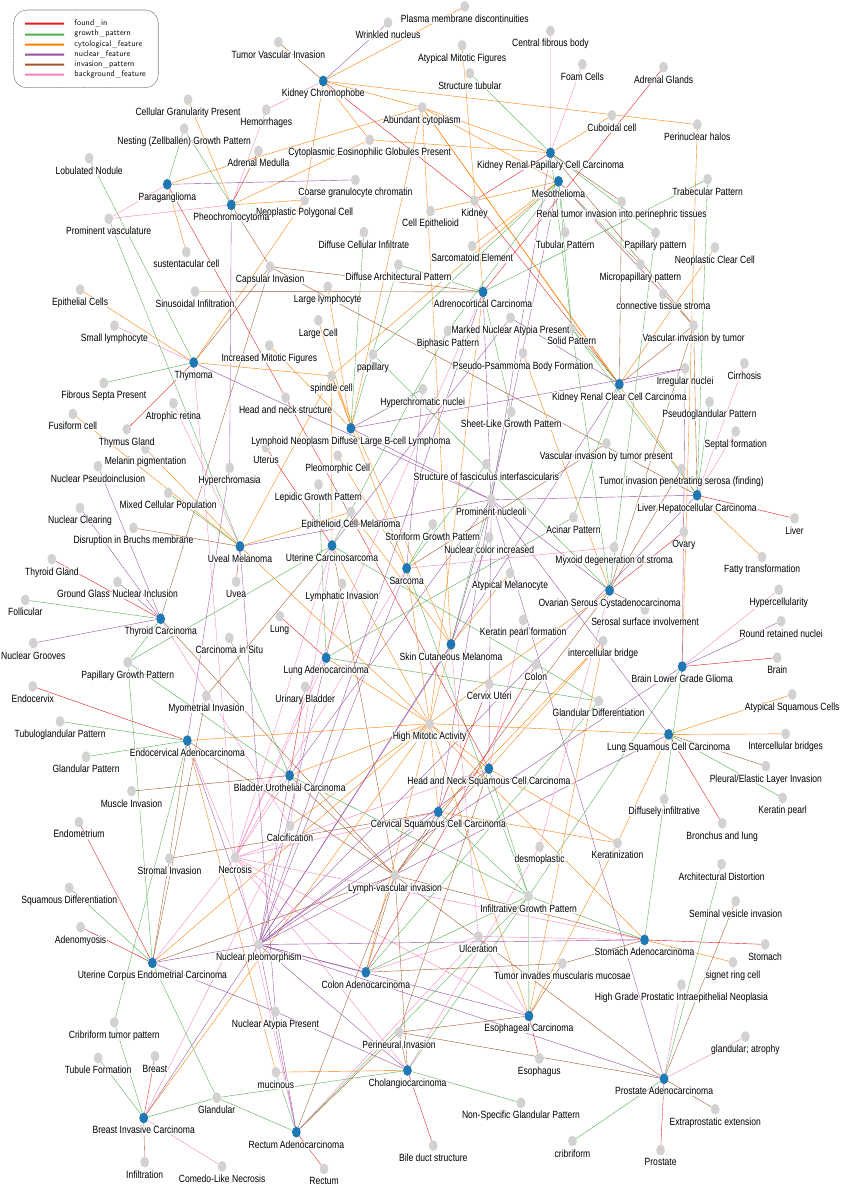}
    \caption{Curated Knowledge Graph}
    \label{fig:placeholder}
\end{figure}

\end{document}

%% file: tables/table5_classifier.tex
\begin{table}[!htbp]
\centering
\color{blue}
\caption{\color{blue}
Ablation of auxiliary histology classification. Combining slide embeddings with primary site improves performance over either input alone.
\textit{Abbreviations: Emb. (Embedding), Acc. (Accuracy), Prec. (Precision), Rec. (Recall), F1 (Macro-F1).}
}
\label{tab:histology_classification_ablation}

\begin{tabular*}{\columnwidth}{@{\extracolsep{\fill}} lllll @{}}
\toprule
Input & Acc. & Prec. & Rec. & F1 \\
\midrule

% Random
% % & 0.03 & 0.03 & 0.03 & 0.03 \\
% & 0.0323 & 0.0323 & 0.0323 & 0.0323 \\

Primary site only
% & 0.84 & 0.68 & 0.74 & 0.69 \\
& 0.8416 & 0.6764 & 0.7393 & 0.6941 \\

Slide emb. only
% & 0.94 & 0.87 & 0.89 & 0.88 \\
& 0.9356 & 0.8685 & 0.8873 & 0.8754 \\

Slide emb. + site
% & \textbf{0.98} & \textbf{0.97} & \textbf{0.98} & \textbf{0.97} \\
& \textbf{0.9752} & \textbf{0.9656} & \textbf{0.9848} & \textbf{0.9711} \\

\bottomrule
\end{tabular*}
\end{table}

%% file: tables/table1.tex
\begin{table*}
\centering
\caption{
Quantitative comparison of existing models on the report generation task of the WSI-Bench \cite{Liang2025wsillava} dataset. \modelname{} outperforms the existing WSI MLLMs in multiple metrics. The results are as reported in WSI-Agents~\cite{lyu2025wsiagents}, and  WSI-LLaVa~\cite{Liang2025wsillava}.
See Sec.~\ref{sec:quantitative results} for detailed analysis. \textit{Abbreviations: B-1/B-2/B-3/B-4 (BLEU-1/2/3/4), R-L (ROUGE-L), M (METEOR), W-P (WSI-Precision).}}\label{tab1}
\begin{tabular}{@{} p{4cm} 
    >{\raggedleft\arraybackslash}p{1.4cm} 
    >{\raggedleft\arraybackslash}p{1.4cm} 
    >{\raggedleft\arraybackslash}p{1.4cm} 
    >{\raggedleft\arraybackslash}p{1.4cm} 
    >{\raggedleft\arraybackslash}p{1.4cm} 
    >{\raggedleft\arraybackslash}p{1.4cm}
    >{\raggedleft\arraybackslash}p{1.4cm}
    >{\raggedleft\arraybackslash}p{1.4cm}@{}}
\toprule
Models &  B-1 & B-2 & B-3 & B-4 & R-L & M & W-P \\
\midrule
GPT-4o~\cite{gpt4o} & 0.202 & 0.069 & 0.030 & 0.016 & 0.132 & 0.167 & 0.067 \\
Quilt-LLaVA~\cite{seyfioglu2024quiltlava} & 0.334 & 0.170 & 0.093 & 0.057 & 0.246 & 0.291 & 0.324  \\
MI-Gen~\cite{chen2024wsicaption} & 0.403 & 0.306 & 0.248 & 0.209 & 0.446 & 0.407 & 0.310 \\
Hist-Gen~\cite{guo2024histgen} & 0.406 & 0.307 & 0.248 & 0.208 & 0.448 & 0.416 & 0.300 \\
WSI-Agents~\cite{lyu2025wsiagents} & 0.444 & 0.308 & 0.231 & 0.183  & 0.448 & 0.472 & 0.440\\
WSI-LLaVA~\cite{Liang2025wsillava} & 0.480 & 0.358 & 0.287 & 0.240 & 0.490 & 0.465 & 0.380\\
\midrule
\textbf{{\modelname}} & \textbf{0.512} & \textbf{0.389} & \textbf{0.316} & \textbf{0.261}  & \textbf{0.519} & \textbf{0.485} & \textbf{0.480} \\
\bottomrule
\end{tabular}
\end{table*}

%% file: tables/table2_kg_ablation.tex
\begin{table*}[!htbp]
\color{blue}
\centering
\caption{\color{blue}
Ablation study of the knowledge pathway. The vision pathway is retained in all conditions, while the information supplied through the knowledge pathway is varied.
% In \modelname{}, the knowledge pathway contains the predicted histology, primary site, and morphological features associated with the predicted histology.
All models are evaluated at the 7B parameter scale. See Sec.~\ref{sec: ablation_studies} for further details.
\textit{Abbreviations: H (predicted histology), PS (primary site), MF (morphological features),
B-1/B-2/B-3/B-4 (BLEU-1/2/3/4), R-L (ROUGE-L), M (METEOR), W-P (WSI-Precision).}
}

\label{tab:knowledge_ablation}

\begin{tabular}{@{} p{4.6cm}
    >{\raggedleft\arraybackslash}p{1.2cm}
    >{\raggedleft\arraybackslash}p{1.2cm}
    >{\raggedleft\arraybackslash}p{1.2cm}
    >{\raggedleft\arraybackslash}p{1.2cm}
    >{\raggedleft\arraybackslash}p{1.2cm}
    >{\raggedleft\arraybackslash}p{1.2cm}
    >{\raggedleft\arraybackslash}p{1.2cm} @{}}
\toprule
Knowledge-path input & B-1 & B-2 & B-3 & B-4 & R-L & M & W-P \\
\midrule

None

& 0.4922 & 0.3745 & 0.3063 & 0.2548 & 0.5185 & 0.4730 & 0.4661 \\

Predicted histology (H)
& 0.5020 & 0.3805 & 0.3088 & 0.2558 & 0.5174 & 0.4789 & 0.4570 \\

% KG features only
% & 0.5095 & 0.3888 & 0.3165 & 0.2623 & 0.4434 & 0.4867 & -- \\

Primary site (PS) only
& 0.4959 & 0.3773 & 0.3072 & 0.2549 & 0.5142 & 0.4750 & 0.4480 \\

\textbf{\modelname{} (H + PS + MF)}
& \textbf{0.5122} & \textbf{0.3885} & \textbf{0.3154} & \textbf{0.2611}
& \textbf{0.5194} & \textbf{0.4849} & \textbf{0.4802} \\

\bottomrule
\end{tabular}
\end{table*}

%% file: tables/table4_param_effec.tex
\begin{table*}[!htbp]
\centering
\caption{\color{blue}
Parameter-efficiency study of the knowledge pathway. Integrating external pathology knowledge enables lower-parameter MLLMs to achieve competitive performance, allowing smaller models to approach the performance of substantially larger models and SOTA systems. Values enclosed in parentheses denote the model parameter size in billions. See Sec.~\ref{sec: ablation_studies} for further evaluation details.
\textit{Abbreviations: B-1/B-2/B-3/B-4 (BLEU-1/2/3/4), R-L (ROUGE-L), M (METEOR), W-P (WSI-Precision), V (Vision), K (Knowledge).}
}
\label{tab:parameter_efficiency}

\begin{tabular}{@{} p{5.4cm} 
    >{\raggedleft\arraybackslash}p{1.2cm} 
    >{\raggedleft\arraybackslash}p{1.2cm} 
    >{\raggedleft\arraybackslash}p{1.2cm} 
    >{\raggedleft\arraybackslash}p{1.2cm} 
    >{\raggedleft\arraybackslash}p{1.2cm}
    >{\raggedleft\arraybackslash}p{1.2cm}
    >{\raggedleft\arraybackslash}p{1.2cm} @{}}
\toprule
Method & B-1 & B-2 & B-3 & B-4 & R-L & M & W-P \\
\midrule
V path only (7B) 
& 0.4922 & 0.3745 & 0.3063 & 0.2548 & 0.5185 & 0.4730 & 0.4661 \\

(V + K) path (1.5B) 
& 0.4792 & 0.3607 & 0.2930 & 0.2426 & 0.5040 & 0.4562 & 0.4290 \\

(V + K) path (3B) 
& 0.4950 & 0.3723 & 0.3011 & 0.2484 & 0.5082 & 0.4719 & 0.4428 \\

(V + K) path (7B)~\textbf{({\modelname})} 
& \textbf{0.5122} & \textbf{0.3885} & \textbf{0.3154} & \textbf{0.2611} 
& \textbf{0.5194} & \textbf{0.4849} & \textbf{0.4802} \\
\bottomrule
\end{tabular}
\end{table*}

%% file: tables/table3_counterfactual.tex
\begin{table}[!htbp]
\centering
\color{blue}
\caption{\color{blue}
Counterfactual knowledge-graph interventions at inference time. We perturb the knowledge supplied to a fixed trained \modelname{} model to evaluate its sensitivity to the retrieved knowledge. The model’s improved report-generation performance with correct knowledge and its degradation under corrupted knowledge provide evidence that it meaningfully utilizes the retrieved knowledge.
\textit{Abbreviation: W-P (WSI-Precision).}
}
\label{tab:counterfactual_kg}

\begin{tabular}{@{} p{5.2cm}
    >{\raggedleft\arraybackslash}p{1.1cm}
    >{\raggedleft\arraybackslash}p{1.1cm}
    >{\raggedleft\arraybackslash}p{1.1cm}
    >{\raggedleft\arraybackslash}p{1.1cm}
    >{\raggedleft\arraybackslash}p{1.1cm}
    >{\raggedleft\arraybackslash}p{1.1cm}
    >{\raggedleft\arraybackslash}p{1.1cm} @{}}
\toprule
Intervention & W-P \\
\midrule
Predicted histology anchor (default)
& 0.4802 \\

Oracle histology anchor
& 0.4830 \\

% Empty knowledge
% & 0.4922 & 0.3745 & 0.3063 & 0.2548 & 0.4349 & 0.4730 & 0.4661 \\

% Confusable-wrong histology anchor
% & 0.2363 & 0.2860 \\

Random-wrong histology anchor
& 0.2540 \\

% Leave-one-node-out
% & -- & -- & -- & -- & -- & -- & -- \\

% Implausible feature insertion
% & -- & -- & -- & -- & -- & -- & -- \\

% Relation-label shuffle
% & 0.5062 & 0.3855 & 0.3141 & 0.2606 & 0.4429 & 0.4835 & 0.4700 \\

\bottomrule
\end{tabular}
\end{table}

%% file: tables/conf_int.tex
\begin{table}[H]
\centering
\caption{
Confidence intervals for \modelname{}'s report-generation metrics, computed over repeated evaluation runs. This indicates the metrics are stable and differences across knowledge path conditions are unlikely to be due to run to run variance alone.
\textit{Abbreviations: B-1/B-2/B-3/B-4 (BLEU-1/2/3/4), R-L (ROUGE-L), M (METEOR), W-P (WSI-Precision).}
}
\label{tab:ci_report_generation}

\begin{tabular}{@{} l
    >{\raggedleft\arraybackslash}p{3.6cm}
    >{\raggedleft\arraybackslash}p{4.8cm} @{}}
\toprule
% Metric & Mean & 95\% CI \\
% \midrule
% B-1 & 0.5124 & (0.5110, 0.5137) \\
% B-2 & 0.3892 & (0.3880, 0.3905) \\
% B-3 & 0.3166 & (0.3154, 0.3177) \\
% B-4 & 0.2625 & (0.2613, 0.2637) \\
% M   & 0.4872 & (0.4844, 0.4899) \\
% R-L & 0.5179 & (0.5166, 0.5192) \\

Metric & 95\% CI \\
\midrule
B-1 & (0.5094, 0.5138) \\
B-2 & (0.3852, 0.3911) \\
B-3 & (0.3122, 0.3185) \\
B-4 & (0.2580, 0.2645) \\
M   & (0.4826, 0.4896) \\
R-L & (0.5179, 0.5203) \\
W-P & (0.4509, 0.4992) \\

\bottomrule
\end{tabular}
\end{table}

%% file: tables/idx52.tex
\begin{table}[H]
%3
\centering
\caption{Qualitative comparison of pathology reports generated by WSI-LLaVA and PathKGen models against the ground truth for a sample whole slide image.}

\small
\setlength{\tabcolsep}{4pt}
\renewcommand{\arraystretch}{1.15}

\begin{tabular}{@{}
p{2.5cm}
p{13cm}
c
c
p{2cm}
@{}}
\toprule

\textbf{Model} &
\textbf{Model Responses}
\\

\midrule
\textbf{WSI}
&
\includegraphics[width=0.3\linewidth]{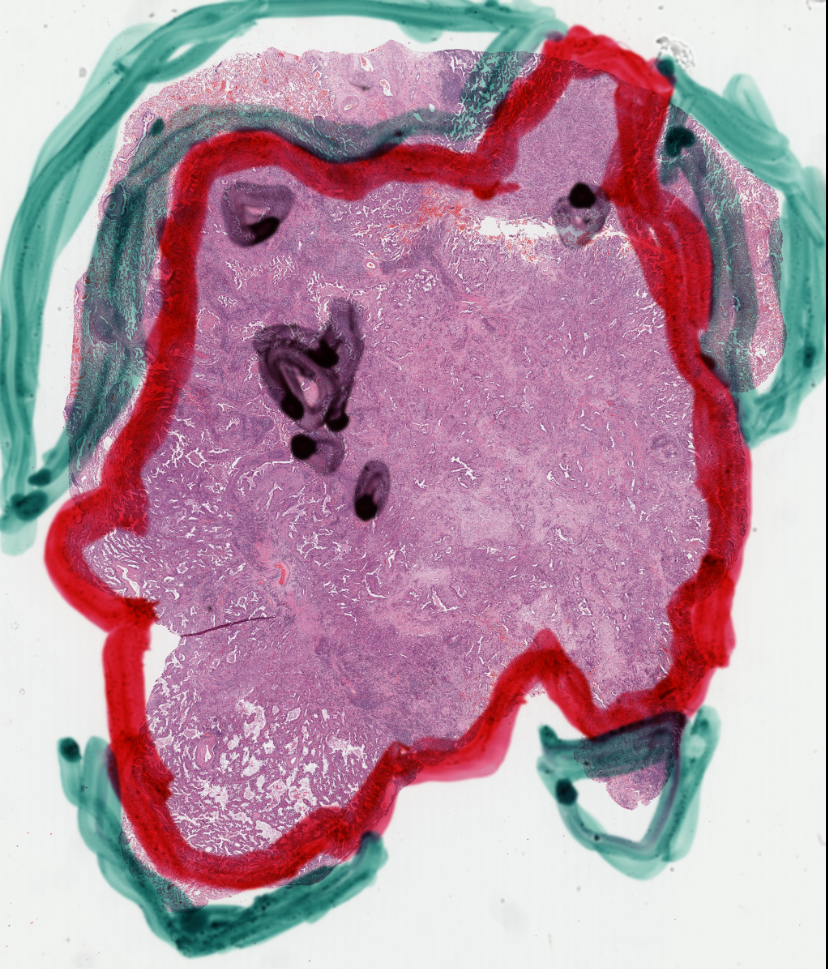}
\\

\midrule

\textbf{Ground Truth}
&
\underline{Microscopic observation of the pathology slide reveals} a presence of tumor cells forming irregular glandular structures that moderately deviate from the normal tissue architecture, consistent with moderate differentiation. The nuclei exhibit slight pleomorphism, with moderate variability in size and shape, \textcolor{Orange}{and there is an intermediate rate of mitosis observed.} Vascular invasion is not specified in the observations available, and there is no mention of necrosis \textcolor{Orange}{or calcification.} \underline{Based on these observations, the diagnosis is} adenocarcinoma, moderately differentiated.
\\

\midrule

\textbf{WSI-LLaVA}
&
\underline{Microscopic observation of the pathology slide reveals} \textcolor{ForestGreen}{a moderately differentiated adenocarcinoma characterized by irregular clusters of epithelial cells} \textcolor{red}{infiltrating the lung tissue.} \textcolor{ForestGreen}{The tumor cells exhibit moderate glandular formation, deviating from normal} \textcolor{red}{alveolar} \textcolor{ForestGreen}{architecture. The nuclei show moderate variability in size and shape, indicative of nuclear pleomorphism.} \underline{There is no evidence of} \textcolor{ForestGreen}{vascular invasion or necrosis} within the observed slide. \underline{Based on these observations, the diagnosis is} \textcolor{red}{adenosquamous carcinoma,} \textcolor{ForestGreen}{moderately-differentiated.}
\\

\midrule

\textbf{PathKGen}
&
\underline{Microscopic observation of the pathology slide reveals} \textcolor{ForestGreen}{a moderately differentiated adenocarcinoma characterized by irregular clusters of epithelial cells} \textcolor{red}{infiltrating the stroma,} \textcolor{ForestGreen}{forming glandular structures that moderately deviate from normal tissue architecture. The nuclei exhibit moderate variability in size and shape, indicative of nuclear pleomorphism, and there is a moderate rate of cell division observed.} \underline{There is no evidence of} \textcolor{ForestGreen}{vascular invasion, necrosis, or calcification within the examined tissue.} \underline{Based on these observations, the diagnosis is} \textcolor{ForestGreen}{adenoc arcinoma, moderately differentiated.}
\\

\bottomrule

\end{tabular}

\vspace{2mm}

\footnotesize
Green: findings present in the ground truth;
Red: hallucinated or contradictory findings;
Orange: ground truth findings omitted by the {\modelname} model;
Underlined: template language.
\label{tab:example_1}
\end{table}

%% file: tables/idx57.tex
\begin{table}[H]
%3
\centering
\caption{Qualitative comparison of pathology reports generated by WSI-LLaVA and PathKGen models against the ground truth for a sample whole slide image.}

\small
\setlength{\tabcolsep}{4pt}
\renewcommand{\arraystretch}{1.15}

\begin{tabular}{@{}
p{2.5cm}
p{13cm}
c
c
p{2cm}
@{}}
\toprule

\textbf{Model} &
\textbf{Model Responses}
\\
\midrule

\textbf{WSI}
&
    \includegraphics[width=0.5\linewidth]{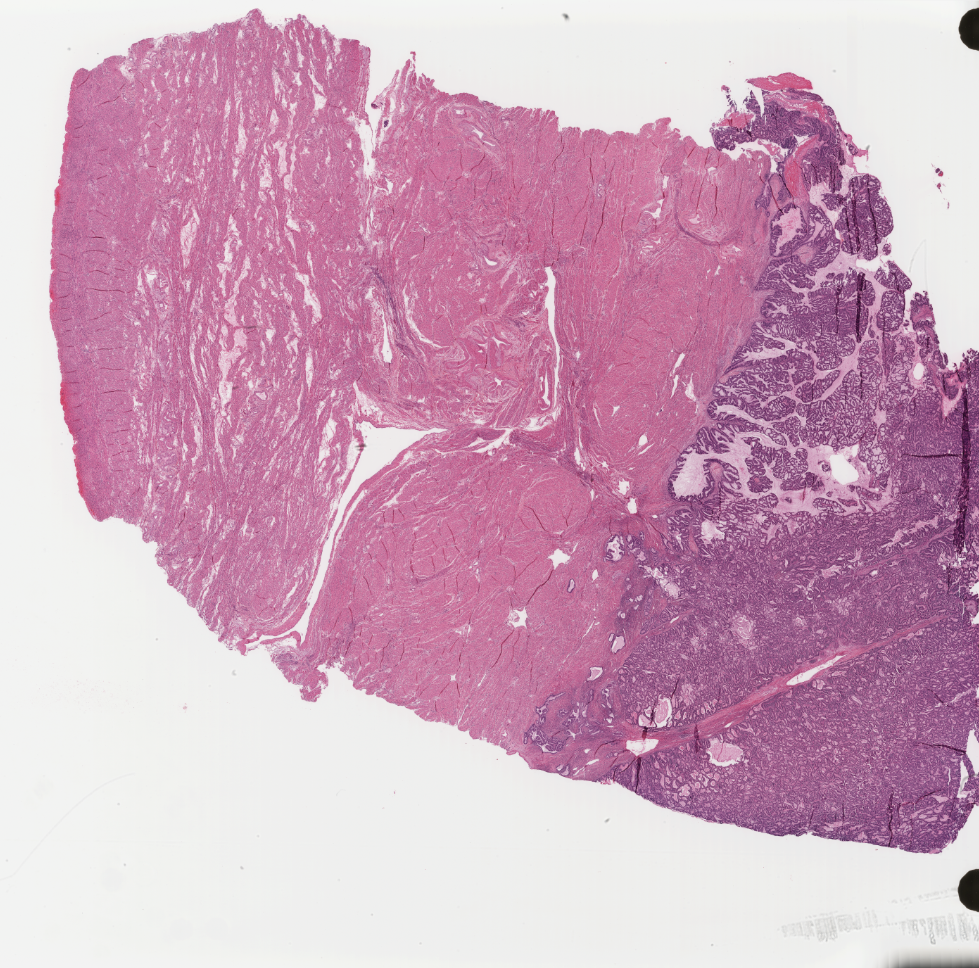}

&
&\\
\midrule

\textbf{Ground Truth}
&
\underline{Microscopic observation of the pathology slide reveals} well-differentiated endometrioid carcinoma cells \textcolor{Orange}{invading minimally into the myometrium.} The architecture is predominantly glandular, exhibiting preserved tubule formation indicative of low-grade differentiation, corresponding to a FIGO Grade 1 classification. There is an absence of lymphovascular invasion, which is a favorable prognostic indicator. The tumor primarily \textcolor{Orange}{shows superficial invasion with no significant features of atypical necrosis or calcification.} \underline{Based on these observations, the diagnosis is} well-differentiated endometrioid adenocarcinoma, FIGO Grade I.
\\

\midrule

\textbf{WSI-LLaVA}
&
\underline{Microscopic observation of the pathology slide reveals} \textcolor{ForestGreen}{well-differentiated glandular structures with minimal deviation from normal endometrial gland architecture, consistent with a FIGO grade 1 classification.} The tumor cells \textcolor{red}{exhibit nuclear features ranging from intermediate to high-grade, indicating a nuclear grade of 2-3.} \underline{There is no evidence of} \textcolor{ForestGreen}{lymphovascular space invasion.} \textcolor{red}{The surrounding non-neoplastic endometrium appears atrophic.} \underline{Based on these observations, the diagnosis is} \textcolor{ForestGreen}{endometrioid adenocarcinoma,} \textcolor{red}{FIGo grade 2.carinoma.}

\\

\midrule

\textbf{PathKGen}
&
\underline{Microscopic observation of the pathology slide reveals} \textcolor{ForestGreen}{endometrioid adenocarcinoma characterized by well-differentiated glandular structures, consistent with a FIGO grade 1 classification.} The tumor cells \textcolor{red}{exhibit minimal nuclear atypia and a low rate of mitotic activity.} \underline{There is no evidence of} \textcolor{ForestGreen}{lymphovascular space invasion,} \textcolor{red}{and the tumor is confined to the endometrium without invasion into the myometrium.} \underline{Based on these observations, the diagnosis is} \textcolor{ForestGreen}{endometrial carcinoma, FIGO Grade 1.}

\\

\bottomrule

\end{tabular}

\vspace{2mm}

\footnotesize
Green: findings present in the ground truth;
Red: hallucinated or contradictory findings;
Orange: ground truth findings omitted by the {\modelname} model;
Underlined: template language.

\label{tab:example_2}
\end{table}

%% file: tables/idx138.tex
\begin{table}[H]
%3
\centering
\caption{Qualitative comparison of pathology reports generated by WSI-LLaVA and PathKGen models against the ground truth for a sample whole slide image.}

\small
\setlength{\tabcolsep}{4pt}
\renewcommand{\arraystretch}{1.15}

\begin{tabular}{@{}
p{2.5cm}
p{13cm}
c
@{}}
\toprule

\textbf{Model} &
\textbf{Model Responses} &
\\
\midrule

\textbf{WSI}
&
    \includegraphics[width=\linewidth]{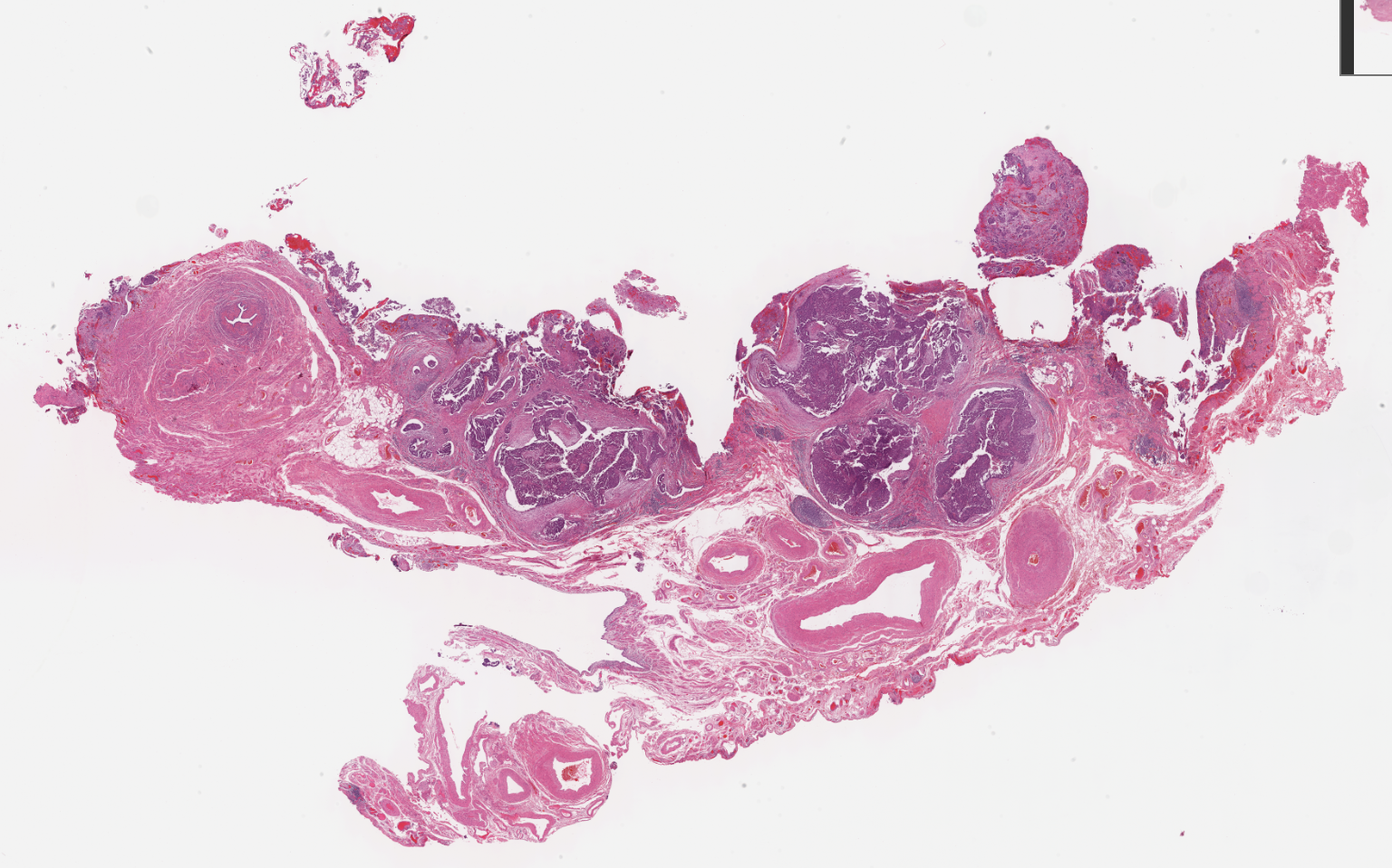}
\\
\midrule

\textbf{Ground Truth}
&
\underline{Microscopic observation of the pathology slide reveals} that the serous adenocarcinoma is characterized by high-grade papillary structures with areas of solid and glandular growth patterns. Tumor cells exhibit significant atypia with prominent nucleoli and high mitotic activity. \textcolor{Orange}{Numerous psammoma bodies}, which are concentrically calcified structures, are present throughout the specimen. There is widespread tumor necrosis and lymphovascular invasion, indicating aggressive behavior of the carcinoma. \textcolor{Orange}{Reactive stromal changes, including fibrosis and secondary inflammation}, are noted, suggesting a chronic inflammatory response. Additionally, foci of calcification are seen, which may correspond to previous necrotic or injured tissues.\underline{Based on these observations, the diagnosis is} high-grade papillary serous carcinoma.
\\

\midrule

\textbf{WSI-LLaVA}
&
\underline{Microscopic observation of the pathology slide reveals} \textcolor{ForestGreen}{a high-grade neoplasm characterized by the presence of malignant epithelial cells forming complex papillary structures} \textcolor{red}{with fibrovascular cores.} \textcolor{ForestGreen}{The tumor cells exhibit marked nuclear atypia, including pleomorphic nuclei with prominent nucleoli} \textcolor{red}{and irregular nuclear contours.} \textcolor{ForestGreen}{There is a high mitotic activity indicative of rapid cell division. Areas of necrosis are present, and there is evidence of vascular invasion within the lymphovascular spaces.} \textcolor{red}{The ovarian stroma shows infiltration by the tumor, and the tumoral cells extend to the serosal surfaces.} \underline{Based on these observations, the diagnosis is} \textcolor{ForestGreen}{high- grade} \textcolor{red}{mullerian-derived} \textcolor{ForestGreen}{adenocarcinoma, papillay type.}
\\

\midrule

\textbf{PathKGen}
&
\underline{Microscopic observation of the pathology slide reveals} \textcolor{ForestGreen}{a high-grade serous carcinoma characterized by papillary and solid growth patterns. The tumor cells exhibit marked nuclear atypia with prominent nucleoli and frequent mitotic figures, indicating a high proliferative activity. There is evidence of lymphovascular invasion, as tumor cells are observed within vascular spaces. Necrosis is present within the tumor tissue,} \textcolor{red}{contributing to its heterogeneous appearance.} \underline{Based on these observations, the diagnosis is} \textcolor{ForestGreen}{serous adenocarcinoma, high grade.}
\\

\bottomrule

\end{tabular}

\vspace{2mm}

\footnotesize
Green: findings present in the ground truth;
Red: hallucinated or contradictory findings;
Orange: ground truth findings omitted by the {\modelname} model;
Underlined: template language.
\label{tab:example_3}
\end{table}